\documentclass[journal]{IEEEtran}

\usepackage[sort]{cite}
\usepackage[pdftex]{graphicx}
\usepackage{xcolor}
\usepackage{tikz}
\usepackage{amsmath}
\usepackage{amsfonts}
\usepackage{bm}
\usepackage[ruled, vlined, commentsnumbered, linesnumbered]{algorithm2e}
\usepackage{algorithmic}
\usepackage{array}
\usepackage{multirow}
\usepackage{url}
\usepackage{hyperref}
\usepackage{tabularx}
\usepackage{booktabs}
\usepackage{kotex}

\usepackage{threeparttable} % tablenotes?
\definecolor{atomictangerine}{rgb}{1.0, 0.6, 0.4}
\definecolor{ballblue}{rgb}{0.13, 0.67, 0.8}
\definecolor{blue-violet}{rgb}{0.54, 0.17, 0.89}
\definecolor{ceruleanblue}{rgb}{0.16, 0.32, 0.75}
\definecolor{coolblack}{rgb}{0.0, 0.18, 0.39}
\definecolor{darkblue}{rgb}{0.0, 0.0, 0.55}
\definecolor{dogwoodrose}{rgb}{0.84, 0.09, 0.41}
\newcommand{\etal}{{\em et al.}}       % et al.
\newcommand{\eg}{{\em e.g.}}           % e.g.
\newcommand{\ie}{{\em i.e.}}           % i.e.
\newcommand{\etc}{{\em etc.}}         % etc.

\newcommand{\HI}[1]{\textcolor{black}{#1}}

\newcommand{\revised}[1]{\textcolor{black}{#1}}
\newcommand{\ejj}[1]{\textcolor{black}{#1}}
\newcommand{\ej}[1]{\textcolor{black}{#1}}

\newcommand{\edit}[1]{\textcolor{black}{#1}}
\newcommand{\rebuttal}[1]{\textcolor{black}{#1}}
\SetKwComment{tcp}{// }{}%
\newcommand{\nonl}{\renewcommand{\nl}{\let\nl\oldnl}}

\begin{document}
% \title{Learning Subject-invariant and Class-relevant\\ Representation via Mutual Information Maximization in BCI}
% \title{Class-relevant Deep Representation Learning\\ via Mutual Information Maximization for BCI}
% \title{Toward Subject Invariant and Class Disentangled Representation in BCI via Cross-Domain Mutual Information Estimator}
% \title{Subject-invariant and Class-relevant Representation\\Learning based on Mutual Information for BCI}
% {Semi-Supervised Deep Adversarial Learning\\for Brain-Computer Interface}
% \title{On Mutual Information Estimation for\\Subject-invariant and Class-relevant\\Deep Representation Learning in BCI}
\title{Mutual Information-driven Subject-invariant and\\Class-relevant Deep Representation Learning in BCI}
% \title{\rebuttal{Mutual Information-driven Subject-invariant\\  Deep Representation Learning in BCI}}
\author{Eunjin~Jeon,~Wonjun~Ko, Jee Seok Yoon, and ~Heung-Il~Suk,~\IEEEmembership{Member,~IEEE}
\thanks{This work was supported by Institute for Information \& Communications Technology Promotion (IITP) grant funded by the Korea government (No. 2017-0-00451, Development of BCI based Brain and Cognitive Computing Technology for Recognizing User's Intentions using Deep Learning). Additional support was provided by Institute of Information \& communications Technology Planning \& Evaluation (IITP) grant funded by the Korea government (MSIT) (No. 2019-0-00079, Department of Artificial Intelligence (Korea University)). (\textit{Corresponding author: Heung-Il Suk})

E. Jeon, W. Ko, and J.S. Yoon are with the Department of Brain and Cognitive Engineering, Korea University, Seoul 02841, Korea (e-mail: eunjinjeon@korea.ac.kr; wjko@korea.ac.kr; wltjr1007@korea.ac.kr). \revised{H.-I. Suk is with the Department of Artificial Intelligence and the Department of Brain and Cognitive Engineering}, Korea University, Seoul 02841 Korea (e-mail: hisuk@korea.ac.kr). }%
}

% \markboth{IEEE TRANSACTIONS ON NEURAL NETWORKS AND LEARNING SYSTEMS}%
\markboth{Under review}%
{Shell \MakeLowercase{\textit{et al.}}: Bare Demo of IEEEtran.cls for IEEE Journals}

\maketitle

\begin{abstract}
In recent years, deep learning-based feature representation methods have shown a promising impact in electroencephalography (EEG)-based brain-computer interface (BCI). \rebuttal{Nonetheless, owing to high intra- and inter-subject variabilities, many studies on decoding EEG were designed in a subject-specific manner by using calibration samples, with no \HI{concern of} its \HI{practical use, hampered by} time-consuming \HI{steps and a large data requirement}. To \HI{this end}, recent studies \HI{adopted a transfer learning strategy, especially domain} adaptation techniques. \HI{Among those, to our knowledge, an adversarial learning has shown its potential in BCIs.}} 
\HI{In the meantime, it is known that adversarial learning-based domain adaptation methods are prone to negative transfer that disrupts learning generalized feature representations, applicable to diverse domains, \eg, subjects or sessions in BCIs. %generalized feature learning interrupt to learn a sufficient feature representation and require the explicit number of subjects for a domain discriminator.}
% Most domain adaptation methods are designed for labeled source and unlabeled target domain whereas BCI tasks generally have multiple annotated domains. 
% (ii) Most of the existing methods do not consider a negative transfer to disrupt generalization ability. 
In this paper, we propose a novel framework that learns class-relevant and subject-invariant feature representations in an information-theoretic manner, without using adversarial learning. To be specific, we devise two operational components in a deep network that explicitly estimate mutual information between feature representations; (1) to decompose features in an intermediate layer into class-relevant and class-irrelevant ones, (2) to enrich class-discriminative feature representation. 
% The  first  component  is  responsible  to  decompose  features  in  an intermediate  layer  into  class-relevant  and  class-irrelevant  ones. The second component operates on top of the class-relevant (referred as `local') features to find more class-discriminative high-level (referred as `global') features, fed into a classifier. While training the network parameters for this second component we impose the mutual information between local features and global features to be maximized in the form of regularization. Our rationale of applying this regularization comes from the possibility of having subject-related information remained in the class-relevant (local) features, even after the decomposition, and the proposed regularization helps extract subject-invariant and class-discriminative features, thereby enhancing classification. 
On two large EEG datasets, we validated the effectiveness of our proposed framework by comparing with several comparative methods in performance. Further, we conducted rigorous analyses by performing an ablation study in regard to the components in our network, explaining our model's decision on input EEG signals via layer-wise relevance propagation, and visualizing the distribution of learned features via t-SNE.} %It is also noteworthy that our method can be applicable to a new subject \rebuttal{without any further training}, thus reducing calibration time for its practical uses. We validate our proposed method on two large motor imagery EEG datasets via comparisons with other competing methods.
\end{abstract}

\begin{IEEEkeywords}
\edit{Brain-Computer Interface; Deep Learning; Electroencephalogram; Motor Imagery; Mutual Information; Transfer Learning; Domain Adaptation; Subject-Independent}
\end{IEEEkeywords}

\IEEEpeerreviewmaketitle

\section{Introduction}
\IEEEPARstart{B}{rain}--computer interface (BCI) allows users to directly communicate or control external devices \edit{based on} thoughts, typically measured \edit{through} electroencephalography (EEG) \cite{graimann2009brain}. EEG signals that measure the electrical activity of \edit{the} brain are usually categorized into two types, \revised{namely,} \emph{evoked} and \emph{spontaneous}, depending on their inducing manner in non-invasive BCIs. Evoked EEGs, \eg, steady-state visually evoked potentials, steady-state somatosensory evoked potentials, and event-related potentials, are derived from immediate automatic responses to an external stimulus regardless of a user's \edit{will.} \edit{In contrast,} spontaneous EEGs induce activation of event-related (de)synchronization (ERD/ERS) when carrying out mental tasks at a user's \edit{will}. Of \edit{the} various \edit{types of EEG signals}, we focus\edit{ed} on \emph{motor imagery} signals showing ERD/ERS induced by \edit{simply imagining body movements.}
% characterized by ERD/ERS induced by imagining body movements without any physical movements. 

\rebuttal{By taking advantage of the controlling system without explicit commands, many studies have \HI{focused} on decoding motor imagery through machine learning. One of the most popular methods is common spatial pattern (CSP) \cite{ramoser2000optimal}. CSP and its variants \cite{ang2008filter,suk2012novel} were employed to design discriminative spatial filters by maximizing differences in their variances between different motor imagery classes to extract features. After conducting CSP \HI{or} its variants, the extracted features were applied to a linear classifier, \eg, linear discriminative analysis (LDA) \cite{vidaurre2010toward}. Recently, deep learning-based methods have drawn increasing attention in BCI researchers by virtue of the possibility to \HI{learn features from data automatically} \cite{lawhern2018eegnet,sakhavi2018learning}. Especially, convolutional neural networks (CNNs) have been well utilized to decode temporal and spatial information of EEG signals and showed remarkable performances \cite{sakhavi2018learning,ko2018deep,schirrmeister2017deep}.}

\begin{figure}[t]
    \centering
    \includegraphics[width=.5\textwidth]{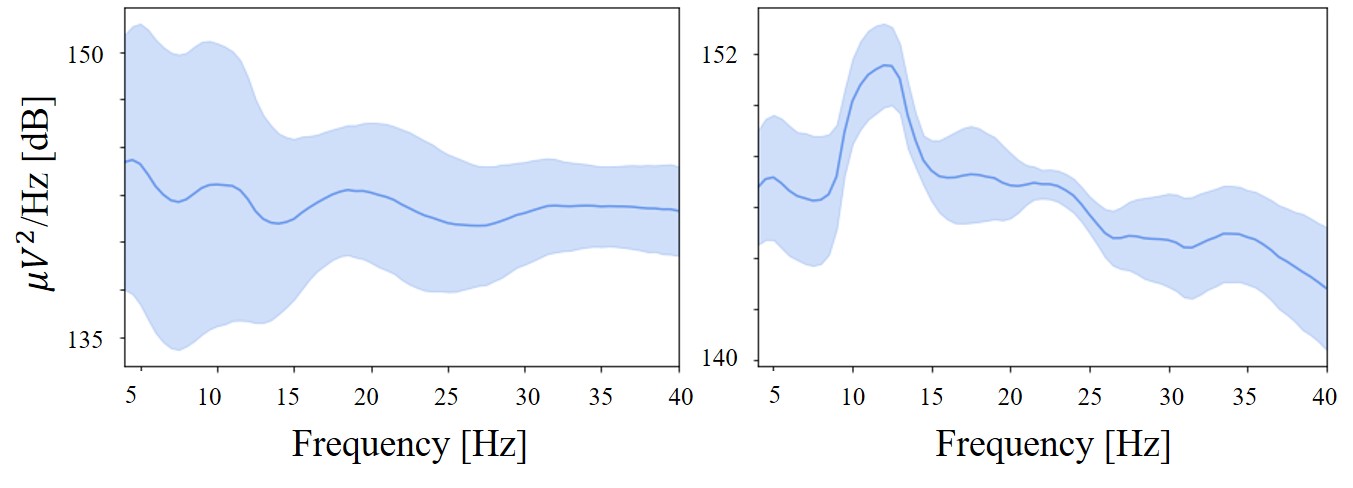}
    % \caption{\edit{(Left) Power spectral density (PSD) curve of subject $45$ in the GIST-motor imagery dataset \cite{cho2017eeg}; subject $45$ has a binary classification accuracy of the common spatial pattern (CSP) as $45.5\%$ and is regarded as a BCI-illiterate subject. (Right) PSD curve of subject $14$ with binary classification accuracy of CSP as $95.0\%$. By taking the average of the trials, we denote the solid line and the area of shaded region as the mean and standard deviation of PSD. While the PSD curve of subject $14$ (right) shows a clear peak in the mu-band, that of subject $45$ does not.}}
    \caption{(Left) Power spectral density (PSD) curve of subject $45$; (Right) PSD curve of subject $14$ in GIST-motor imagery dataset \cite{cho2017eeg}. By taking the average of the trials, we denote the solid line and the area of shaded region as the mean and standard deviation of PSD. Both subjects show different patterns in PSD, which can be regarded as a domain shift.}
    \label{fig:psd_curve}
\end{figure}

However, the motor imagery EEG shows high variability among subjects (inter-subject) and sessions for the same subject (intra-subject) \cite{jayaram2016transfer} on account of inherent background neural activities, fatigue, concentration levels, \etc~ Fig. \ref{fig:psd_curve} presents the power spectral density (PSD) curves of motor imagery EEG signals obtained from two subjects. Both curves are plotted using all motor imagery EEG samples over the sensorimotor area. In Fig. \ref{fig:psd_curve}, the right panel presents a clear peak in the mu-band with relatively small variations (blue shading) among samples, whereas the left panel shows large variations without any clear pattern. \rebuttal{\HI{Because of those unpredictable high variations and less clearly observable patterns inherent in EEG signals,} it is challenging to train a subject-invariant model, which is applicable to different datasets or subjects. In this regard, an EEG decoding model trained on one subject causes performance degradation when applied to other subjects \cite{pan2009survey,chai2016unsupervised,jeon2019domain}. Therefore, training a model for each subject is a typical approach to decode brain signals despite time-consuming and amount of data-requirable process \cite{lin2017improving}. In order to address the limitation, previous studies exploited multiple subjects and/or sessions data simultaneously to train \HI{their respective models} through transfer learning \cite{jayaram2016transfer,lin2017improving}.}

\rebuttal{We focus on, in this work, boosting model generalization among subjects in the transfer learning manner \cite{jayaram2016transfer,ozdenizci2020learning}, by considering a subject as a domain \cite{pan2009survey}. Owing to the unfavorable property of motor imagery EEG, \ie, high inter- and intra-subject variabilities \cite{jayaram2016transfer}, a large distributional discrepancy was observed on \HI{a feature space among} different subjects or sessions, referred to as a \emph{domain shift} \cite{chai2016unsupervised,jeon2019domain}. Hence, mitigating the domain shift is one of the \HI{important objectives} in the transfer learning for BCI tasks \cite{ozdenizci2019transfer,jeon2019domain,ozdenizci2020learning}.}

% Regarding the domain shift problem, numerous studies on machine learning, referring to domain adaptation, have been conducted \cite{ganin2016domain,wang2018deep,tzeng2017adversarial}. However, a direct application of domain adaptation techniques to BCI tasks is challenging because BCI tasks generally exploit multiple annotated domains, \ie, subjects; the previous domain adaptation methods most considered two domains: labeled \emph{source} and unlabeled \emph{target} domain \cite{ganin2016domain,tzeng2017adversarial}.

Regarding the domain shift problem, numerous studies on machine learning, referring to domain adaptation, have been conducted \cite{ganin2016domain,wang2018deep,tzeng2017adversarial}. \rebuttal{Particularly, most methods \HI{tried to tackle the problem by introducing a domain discriminator in their framework to discover features indistinguishable among domains} via adversarial training like a generative adversarial network \cite{ganin2016domain}. \HI{In the same line of work, \cite{ozdenizci2019transfer,ozdenizci2020learning} proposed a domain adversarial network to learn a subject-invariant representation for motor imagery classification in BCIs, which is applicable for more than two subjects by devising a subject or domain discriminator.}} 

\rebuttal{However, recent studies have pointed two \HI{major} limitations of those approaches: \HI{First, an adversarial} learning, which induces to alleviate a domain shift, is \HI{likely to disrupt} feature representation learning, due to \HI{restraint of} domain-specific \HI{variations,} while not considering class-related distribution across domains \cite{liu2019transferable}. \HI{Second, most} domain adaptation methods may not only improve the feature representation of a target domain, but also corrupt it. This corrupting phenomenon is called \emph{negative transfer} \cite{peng2019domain}. In this regard, Peng \etal~\cite{peng2019domain} proposed \HI{a framework, where} the source domain was sorted as \emph{domain-invariant}, \emph{domain-specific}, and \emph{class-irrelevant} to alleviate \HI{the effect of} negative transfer. To this end, it is of great importance to differentiate positive and negative transferable factors from data \HI{of various domains}.}

\HI{In this work, we propose a novel framework to learn subject-invariant feature representations for motor imagery EEG signals classification in an information-theoretic manner, instead of using the adversarial learning strategy, towards subject-transferable learning and ultimately subject-independent BCIs. Concisely, we introduce a specially designed network component that decomposes feature maps of an intermediate layer in a feature extractor into class-irrelevant and class-relevant ones via mutual information estimation. Further, owing to the possibility of having subject-related information remained in the class-relevant features that could be negative transfer, even after the decomposition, we further devise a regularization mechanism that imposes the extracted class-relevant (regarded as `local') features and the next-level of abstract (regarded as `global') features to have maximal mutual information. The rationale of imposing this regularization of maximal mutual information between local and global features is as follows. As the global feature representation layer is closer to a classifier, it is more likely to receive class-discriminative information in backpropagation from the classification loss. By imposing such information to be reflected in the local features by means of mutual-information maximization, it is expected to discard the remaining subject-related information in the local features, but to keep the class-relevant features only. Thus, it is expected to lessen the effect of negative transfer and to enhance the discriminative power of learned feature representations. In other words, to jointly consider domain shifts between multiple domains (\ie, subjects) and to estimate mutual information from them, we devise a novel network architecture to estimate mutual information in high- and low-level representations in a subject-independent manner. As a result, our trained model can be subject-invariant and suitable for applying to new subjects in a way of zero training, \ie, no subject-dependent adaptation or calibration is required.}

\HI{Compared to the existing domain-adaptation work in BCIs \cite{ozdenizci2019transfer,ozdenizci2020learning}, our method is less concerned on the negative transfer thanks to the mutual information-driven learning. In addition, unlike those existing works that use a domain discriminator that identifies a subject's id with regard to an input EEG signals, our framework does not use a subject's id during training. In the circumstance of incrementally adding samples of new training subjects, their architectures need to accommodate the increasing number of domains, \ie, training subjects, and modify their classifier accordingly. However, since our proposed method does not use the information of subjects' id, there is no need to revise or modify our network, thus being scalable to domains or training subjects.}

\HI{We evaluated our proposed framework on GIST \cite{cho2017eeg} and KU \cite{lee2019eeg} motor imagery datasets.} %for classification to decide the intention of left- or right-hand movements. 
\rebuttal{Our experimental results \HI{demonstrated} that (i) the \HI{feature decomposition into class-relevant and class-irrelevant helped enhance the performance by diminishing distributional difference in features among} subjects and (ii) maximizing mutual information between high- and low-level \HI{representations encouraged to enrich class-discriminative features and to} boost the performance.} Our results showed promising performance compared to the competing methods trained with data from multiple subjects. 
The main contributions of our work are three-fold:
\begin{itemize}
    \item \rebuttal{First, we propose a novel \HI{deep-learning framework that learns subject-invariant and class-relevant feature representations in an information-theoretic and end-to-end manner.}}
    % \item \rebuttal{Our proposed network showed plausible results of classification performance for an unseen subject in terms of zero-shot approach.}
    \item \HI{Our proposed components of feature decomposition and feature enrichment can be naturally plugged into the existing network architectures, \eg, EEGNet \cite{lawhern2018eegnet} and Deep ConvNet \cite{schirrmeister2017deep}}.
    \item \HI{On two large motor imagery EEG datasets, our method achieved the best performance in both cross-subject learning and zero-training scenarios.}
    % \item \HI{Our proposed network can be applied to \HI{unseen subjects without any fine-tuning or calibration steps involved, thereby moving towards zero-training in BCIs.}}
    % \item \rebuttal{Finally, we analyze our proposed network from the neurophysiological viewpoints and then showed that class-decomposition did not deteriorate the neurophysiological representational power.}
\end{itemize}

\begin{figure*}[t]
\centering
\includegraphics[width=0.85\textwidth]{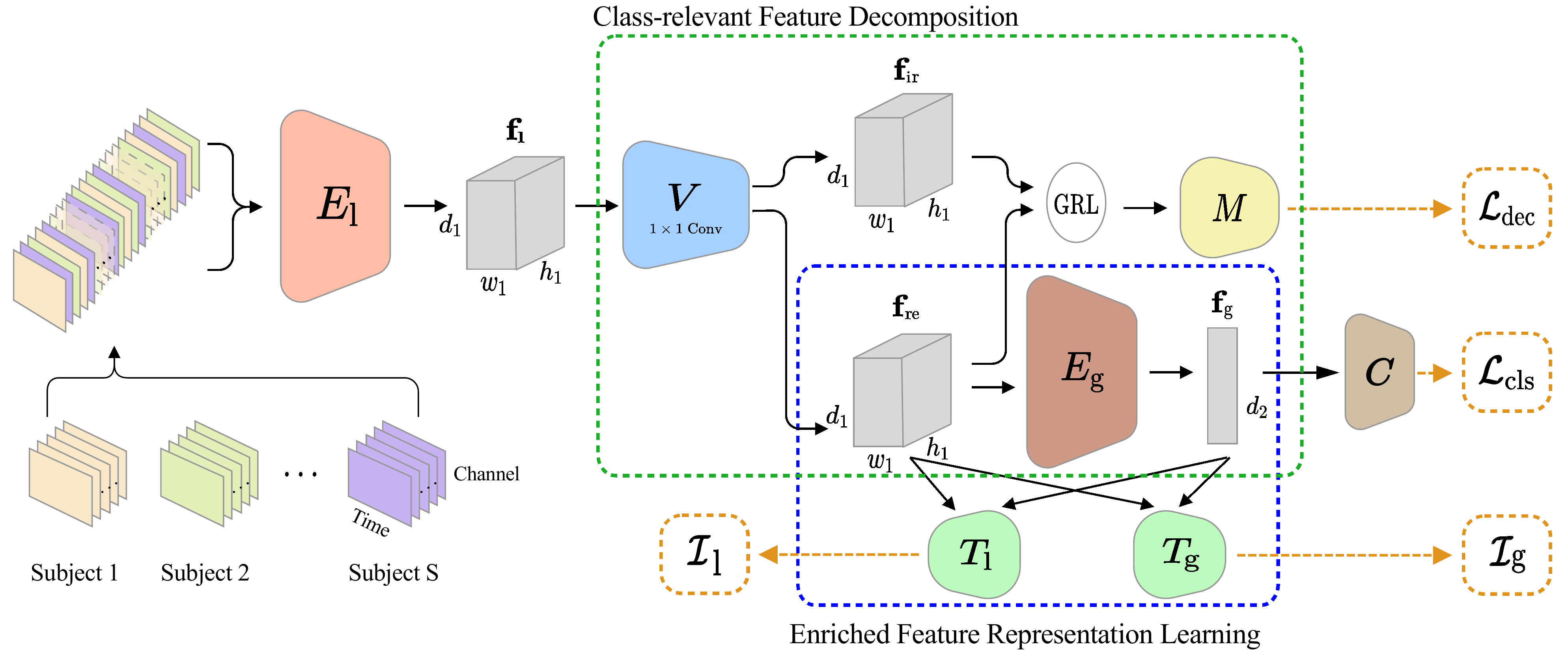}
\caption{\rebuttal{Overview of the proposed network. We randomly select trials regardless of the subjects for a mini-batch. \HI{After mapping an input $\mathbf{x}$ to a local feature $\mathbf{f}_\text{l}$ by a local encoder $E_\text{l}$}, a point-wise convolutional layer $V$ embeds it into a class-relevant feature $\mathbf{f}_\text{re}$ and a class-irrelevant feature $\mathbf{f}_\text{ir}$ \HI{with the help of mutual information estimator $M$}. Subsequently, a global encoder $E_\text{g}$ \HI{operates on top of} the class-relevant feature, and then extracts global feature $\mathbf{f}_\text{g}$.} A classifier $C$ takes the global feature $\mathbf{f}_\text{g}$ to \HI{identify the class label of an input EEG}. \rebuttal{Furthermore, to enhance the representational power of the global feature, we maximize mutual information between two features \HI{of $\mathbf{f}_\text{re}$ and $\mathbf{f}_\text{g}$} by using two networks, \ie, $T_\text{l}$ and $T_\text{g}$.}}
% \caption{Overview of the proposed network. We randomly select trials regardless of \edit{the} subjects for a mini-batch. After \edit{local} encoder $E_l$ maps \edit{input} $\mathbf{x}$ to \edit{local} feature $\mathbf{f}_l$, \edit{global} encoder $E_g$ receives the local feature, \edit{and then} extracts \edit{global} feature $\mathbf{f}_g$. To embed \edit{a} subject-invariant feature, we maximize mutual information between two features by using \edit{local} discriminator $T_{l}$ and global discriminator $T_{g}$. We split \edit{global} feature $\mathbf{f}_g$ into \emph{class-relevant} feature $\mathbf{f}_{re}$ and \emph{class-irrelevant} feature $\mathbf{f}_{ir}$. Then, the classifier $C$ takes $\mathbf{f}_{re}$ and discriminate the corresponding class. Through a gradient reversal layer (GRL), we reduce \edit{the} mutual information between $\mathbf{f}_{re}$ and $\mathbf{f}_{ir}$ estimated from \edit{mutual} information neural estimator $M$. Finally, \edit{decoder} $D$ \edit{was used to reconstruct} the original input from \edit{concatenated} feature $\hat{\mathbf{f}}_g$.}
\label{fig:overall}
\end{figure*}

\section{Related Work}

\subsection{Transfer Learning in BCI}
In general, many methods for decoding EEGs are devised for individual use with independent training per subject; this is because of high inter- and intra-subject variabilities. However, the calibration and training of the decoding model requires plenty of data and is time-consuming \cite{lin2017improving,jayaram2016transfer}. To cope with the aforementioned limitation, various studies have focused on training the decoding model with multiple subjects and/or sessions with the objective of learning a subject-invariant representation. These approaches are divided into two: (i) zero- or few-shot learning for decoding training data of unseen subject \cite{ozdenizci2019transfer,kindermans2014integrating,azab2019weighted} and (ii) performance improvement by incorporating data of other subjects \cite{lin2017improving,wei2018subject,jeon2019domain}. In particular, deep learning-based methods \cite{chai2016unsupervised,jeon2019domain,ozdenizci2019transfer,ozdenizci2020learning} utilize a domain adaptation approach to deal with these issues. They all constrain the distributional discrepancy between subjects by minimizing the maximum mean discrepancy in the latent space \cite{chai2016unsupervised} or making the encoder to confuse the domain \cite{jeon2019domain,ozdenizci2019transfer,ozdenizci2020learning}. Among those studies, two proposed methods were fundamentally devised for only two subjects \cite{chai2016unsupervised,jeon2019domain} and did not utilize diverse subjects. In \cite{ozdenizci2019transfer}, a classifier was trained from the pre-trained subject-invariant representation, and hence, confusing the domain and identifying the class were not trained in an end-to-end manner. \rebuttal{Although \cite{ozdenizci2020learning} proposed a method to extract a subject-invariant and class-discriminative features among multiple subjects in an end-to-end manner, there still remain some challenges. Since \cite{ozdenizci2020learning} trains a domain discriminator by explicitly using subjects' id (or domain labels), \HI{when samples of new training subjects are added incrementally}, it is required to modify the classifier architecture, thus to increase tunable parameters. However, our proposed network achieves domain adaptation among multiple subjects and classification in an end-to-end manner by estimating mutual information, without explicitly using subjects' id. In this regard, our proposed network is scalable to domains. Further, as no adversarial learning mechanism is involved, it does not have a risk of experiencing negative transfer to deteriorate the performance\HI{, caused by the adversarial learning.} \cite{liu2019transferable}.}

\subsection{Transfer Learning in Machine Learning}
Various studies have been done to mitigate differences between \emph{source} and \emph{target} domains in domain adaptation, \ie, a case of transfer learning. \edit{In \cite{wang2018deep}, the domain adaptation was categorized into \emph{one-} and \emph{multi-step}} approaches depending on the presence of an intermediate domain, decreases the gap between \emph{source} and \emph{target} domains. The one-step domain adaptation was achieved by means of a domain discriminator, which guides its features to become indistinguishable between domains by reversing its gradient during training through \ejj{the} adversarial learning \cite{ganin2016domain,tzeng2017adversarial}. 
% \todo{However, they considered a single \emph{target} domain by using one \emph{source} domain. In the case of multiple \emph{source} domains, multiple domain discriminators identify the domain between a \emph{target} domain and each source domain, separately \cite{zhao2018adversarial,xu2018deep}. As we should denote a specific \emph{target} domain to take advantage of existing methods, they are inapplicable for our task dealing with several \emph{source} domains.}
\rebuttal{However, Liu \etal~\cite{liu2019transferable} showed that adversarial feature learning can potentially damage their original feature representation from a viewpoint of transfer learning.} \HI{In this paper, we take a different strategy and propose a novel framework to learn domain adaptation by means of  mutual information estimation to lessen} the distributional discrepancy among domains. For this, we assume that a latent space maximized via the mutual information between multiple domains can be viewed as a commonly shared space among those domains. In addition, recently, studies have focused on disentangled representation learning in terms of transfer learning to prevent \emph{negative transfer} \cite{peng2019domain,wang2019transferable}. Peng \etal~\cite{peng2019domain} divided a latent representation into \emph{domain-invariant}, \emph{domain-specific}, and \emph{class-irrelevant} features by minimizing mutual information among them for domain-agnostic learning. \rebuttal{Similar to those approaches, we decompose our feature representations into \emph{class-relevant} and \emph{class-irrelevant} features and then boost the class-relevant feature to contain more \emph{domain-invariant} (\ie, \emph{subject-invariant}) information.} In particular, we utilize mutual information between two feature representations to ensure decomposition.

\section{Methods}
In this work, we regard each subject as one domain. Thus, we assume that for $S$ subjects, $\mathcal{D}^{s}=\{(\mathbf{x}_s^{(i)},\mathbf{y}_s^{(i)})\}_{i=1}^{n_s}$ with $n_s$ labeled samples, where $s\in\{1\dots,S\}$.  \rebuttal{Let $\mathbf{x}_s^{(i)}\in\mathbb{R}^{n_c\times n_t}$ denote a raw EEG trial including spatio-temporal information, where $n_c$ and $n_t$ are the \HI{number} of electrode channels and timepoints, respectively.} We also define $\mathbf{y}_s^{(i)}\in\{0,1\}^2$ as the corresponding class label, \ie, left- and right-hand. Hereafter, for uncluttered, we omit the superscripts ($s$ and $i$) without loss of generality.

The goal of this work is to build a deep neural network \emph{robustly} applicable for multiple subjects. In other words, we develop an intention identification system that can be generally applicable for all \HI{subjects}. Further, \HI{our trained model} \rebuttal{can be applied to a new subject's \HI{data} without any further \HI{calibration and/or training steps, thus towards zero-training}.} 

\rebuttal{An overall framework of our proposed method is shown in Fig. \ref{fig:overall}. Our network first discovers a feature representation including spatio-temporal information through existing \HI{CNN-based networks}. However, to filter \HI{out class-irrelevant representation}, we utilize an intermediate feature representation, \ie, an output of a local encoder $E_\text{l}$ (local feature $\mathbf{f}_\text{l}$). After the local feature $\mathbf{f}_\text{l}$ is decomposed into \emph{class-relevant} feature $\mathbf{f}_\text{re}$ and \emph{class-irrelevant} feature $\mathbf{f}_\text{ir}$, only class-relevant feature $\mathbf{f}_\text{re}$ \HI{passes through the following layers of a global encoder $\mathbf{f}_\text{g}$) and a classifier $C$ to identify} the corresponding class of \HI{an} input $\mathbf{x}$. To help our network diminish dependency between two characteristics in features, \ie, \emph{class-relevant} feature and \emph{class-irrelevant} feature, we leverage a \HI{neural} mutual information estimator $M$ (green box in Fig. \ref{fig:overall}). In the meantime, another two mutual information estimators ($T_\text{l}$ and $T_\text{g}$) are utilized to learn more \emph{subject-invariant} and class-discriminative feature \HI{representation} by maximizing mutual information between the class-relevant feature and the global feature among multiple subjects (blue box in Fig. \ref{fig:overall}).}
% An overall framework of our proposed method is shown in Fig. \ref{fig1}. \revised{Our network first discovers a feature representation that maximize\edit{s} mutual information among subjects (blue part in Fig. \ref{fig1}) and then decomposes the feature representation into the \emph{class-relevant} and \emph{class-irrelevant} features (yellow part in Fig. \ref{fig1}).} \edit{Our} proposed network has an autoencoder structure. In the encoding-related block (blue box in Fig. \ref{fig1}), \edit{$\mathbf{x}$ is input into} the \emph{subject-invariant} feature space \edit{by} encoder $E$, \ie, $E_l$ and $E_g$. \edit{The decoding-related block (yellow box in Fig. \ref{fig1}) comprises three components:} (i) \edit{classifier} $C$ to identify the class of input $\mathbf{x}$; (ii) mutual information estimator $M$ to help our network diminish dependency among two characteristics in features, \ie, \emph{class-relevant} feature $\mathbf{f}_{re}$ and \emph{class-irrelevant} feature $\mathbf{f}_{ir}$; and (iii) decoder $D$ to reconstruct input $\mathbf{x}$ with concatenated feature $\hat{\mathbf{f}}_g$, \ie, [$\mathbf{f}_{re}$, $\mathbf{f}_{ir}$]. Later, the trained network is fine-tuned for a new subject by \edit{only} using his/her data.
\subsection{Class-relevant Feature Decomposition}
\rebuttal{As aforementioned, all feature representations are not necessarily useful for classification in terms of transfer learning \cite{peng2019domain}. Hence, we assume that a feature representation can be decomposed into \emph{class-relevant} and \emph{class-irrelevant} factors and then introduce a method to separate a \emph{class-relevant} feature out by minimizing mutual information between two factors.} 

\rebuttal{First, our encoder $E$ is comprised of the local encoder and the global encoder. The local encoder, $E_\text{l}$, maps input $\mathbf{x}$ onto the local feature $\mathbf{f}_\text{l}=E_\text{l}(\mathbf{x}) \in \mathbb{R}^{h_1\times w_1\times d_1}$, where $h_1$, $w_1$, and $d_1$ represent the height, width, and depth of the feature, respectively. The global encoder $E_\text{g}$ computes the global feature $\mathbf{f}_\text{g}$ that is fed into a classifier $C$. In other words, if the local encoder embeds the local feature, the remaining layers are regarded as the global encoder. The encoder $E$ is structured such that it produces an intermediate local feature and then the global feature.}

\rebuttal{Before feature decomposition in the local feature $\mathbf{f}_\text{l}$, we apply a point-wise convolution $V$ to the local feature considering cross-feature map correlation \cite{chollet2017xception} to alleviate unexpected information loss. Subsequently, we obtain two features; one is related to class-relevant feature $\mathbf{f}_\text{re}$ among subjects and the other one is related to class-irrelevant feature $\mathbf{f}_\text{ir}$ that can be regarded as subject-specific factors regardless of the classification, where $\mathbf{f}_\text{re}, \mathbf{f}_\text{ir}\in\mathbb{R}^{h_1\times w_1\times d_1}$. The global encoder $E_\text{g}$ takes only the class-relevant feature as an input and then outputs a global feature $\mathbf{f}_\text{g}$. The classifier $C$ is trained by minimizing the softmax cross-entropy loss}:
\begin{equation}\label{eq:l_cls}
    \mathcal{L}_{\text{cls}} =  -\sum_{i=1}^{N}\mathbf{y}^{(i)}\log(C(E_\text{g}(\mathbf{f}_\text{re}^{(i)})))
\end{equation}
where $\mathbf{y}^{(i)}$ denotes a one-hot label vector for an input $\mathbf{x}^{(i)}$ \rebuttal{and $N$ is the number of samples in a mini-batch}. \rebuttal{The class-relevant factors are sufficient for the final classification task after embedding to the global feature through the global encoder.}

% we split global feature $\mathbf{f}_g\in\mathbb{R}^{d_2}$ into two parts such that one part is related to class-relevant feature $\mathbf{f}_{re}$ and the other part is related to class-irrelevant feature $\mathbf{f}_{ir}$, where $\mathbf{f}_{re}, \mathbf{f}_{ir}\in\mathbb{R}^{d_2/2}$. Then, classifier $C$ is trained by minimizing the softmax cross-entropy loss with only the class-relevant feature as input:
% \begin{equation}\label{eq:l_cls}
%     \mathcal{L}_{\text{cls}} =  -\sum_{i=1}^{N}\mathbf{y}^{(i)}\log(C(\mathbf{f}_{re}^{(i)}))
% \end{equation}
% where $\mathbf{y}^{(i)}$ denotes an one-hot label vector for input $\mathbf{x}^{(i)}$. 

\rebuttal{\HI{In order to better} decompose local feature $\mathbf{f}_\text{l}$ into two factors, \ie, $\mathbf{f}_\text{re}$ and $\mathbf{f}_\text{ir}$, we exploit mutual information between them,} which is defined in a form of the Kullback-Leibler (KL) divergence between the their joint distribution $\mathbb{J}$ and the product of marginals of them $\mathbb{M}$. Recently, mutual information neural estimation (MINE) \cite{belghazi2018mutual} showed that \rebuttal{mutual information is estimated based on deep learning by reformulating} it as the Donsker-Varadhan representation \cite{donsker1983asymptotic}: 
\begin{align}\label{eq:mine_kl}
    \mathcal{I}_{\Theta}(X;Y)&=D_{\text{KL}}(\mathbb{J}||\mathbb{M})\notag\\ &= \sup_{\theta\in\Theta}\mathbb{E}_{\mathbb{J}}[T_{\theta}(x,y)]-\log(\mathbb{E}_{\mathbb{M}}[e^{T_\theta(x,y)}])
\end{align}
where $T$ is a neural network parameterized by $\theta$. Here, the product of marginal distributions $\mathbb{M}$ is induced by shuffling the samples from the joint distribution along the batch axis. MINE \cite{belghazi2018mutual} achieves the estimation of mutual information between continuous random variables by maximizing Eq. \eqref{eq:mine_kl}.

\rebuttal{However, as we only focus on maximizing the mutual information, the exact value of mutual information in terms of training the neural network is not required. In this regard, Deep infomax (DIM) \cite{hjelm2018learning} takes advantage of the Jenshen-Shannon (JS) divergence as an alternative of Eq. \eqref{eq:mine_kl} by following \cite{nowozin2016f}. Thus, Eq. \eqref{eq:mine_kl} can be rewritten as follows: 
\begin{align}\label{eq:mine_JSD}
    \HI{\mathcal{I}}_\Theta(X;Y) = \mathbb{E}_{\mathbb{J}}[-\text{sp}(-T_\theta(x,y))]-\mathbb{E}_{\mathbb{M}}[\text{sp}(T_\theta(x,y))]
\end{align}
where $\text{sp}(z)=\log(1+e^z)$ is \HI{a} softplus function.}

\rebuttal{From the viewpoint of the feature decomposition,} we make use of mutual information between $\mathbf{f}_\text{ir}$ and $\mathbf{f}_\text{re}$. \HI{In other words, by introducing a network $M$ to estimate mutual information between $\mathbf{f}_\text{ir}$ and $\mathbf{f}_\text{re}$ and penalizing a large mutual information value of $M$, we make the two factors of $\mathbf{f}_\text{ir}$ and $\mathbf{f}_\text{re}$ to be decomposed with minimal information overlap. %ensures our network to better decompose local feature $\mathbf{f}_\text{l}$ into two factors,} $\mathbf{f}_\text{ir}$ and \HI{$\mathbf{f}_\text{re}$}.
}
Hence, \HI{the neural estimator} \rebuttal{$M$ is trained with the \ejj{the local encoder $E_\text{l}$ and the point-wise convolutional layer $V$} \HI{based on the output of a neural estimator $M$, \ie, $\mathcal{I}_\text{d}(\mathbf{f}_{re}^{(i)};\mathbf{f}_{ir}^{(i)})$, which defines the} decomposition loss $\mathcal{L}_\text{dec}$}:
% $M$, is introduced and then trained by mutual information $\mathcal{I}_d$, denoted as: 
\begin{equation}\label{eq:mine}
    \mathcal{L}_\text{dec} = \min_{E_\text{l},V}\max_{M}\sum_{i=1}^{N}\mathcal{I}_\text{d}(\mathbf{f}_{re}^{(i)}\HI{;}\mathbf{f}_{ir}^{(i)}).
\end{equation}
% \rebuttal{where $\mathcal{I}_\text{d}$ denotes mutual information defined as a form of Eq. \eqref{eq:mine_JSD}.}
In \cite{belghazi2018mutual}, the parameters of MINE \cite{belghazi2018mutual} are trained by \emph{gradient ascent} owing to the use of the Donsker-Varadhan representation \cite{donsker1983asymptotic}. However, in our case, as we want to minimize the mutual information between $\mathbf{f}_\text{re}$ and $\mathbf{f}_\text{ir}$, we apply a gradient reversal layer (GRL) \cite{ganin2016domain}. The GRL passes features innate to $M$ in forward propagation, but reverses the gradients in backpropagation, \ie, reversed gradients, during training.

\subsection{\rebuttal{Enriched Feature Representation Learning}} 
\rebuttal{Meanwhile, in order to generalize our proposed method in the subject-independent manner, it is of great importance to learn a subject-invariant and class-discriminative feature. However, although we separate the class-relevant feature and the class-irrelevant feature by the means of MINE $M$ \cite{belghazi2018mutual}, the class-relevant feature can potentially lose much of the class-related information or still contain subject-specific information.} 

\HI{In this regard, we further introduce another mutual-information-oriented mechanism that helps enrich the discriminative power of the feature representation $\mathbf{f}_{\text{g}}$, which is fed into a classifier. Specifically, based on deep infomax (DIM)  \cite{hjelm2018learning}, we devise two sub-networks, denoted as $T_{\text{l}}$ and  $T_{\text{g}}$ in Fig. 2, to maximize the mutual information between two features of the class-relevant feature $\mathbf{f}_\text{re}$ and the global feature $\mathbf{f}_{\text{g}}$.} 
%To this end, we exploit \HI{two sub-networks}, \ie, DIM by maximizing the mutual information between the input and encoder's output to learn a useful feature representation \cite{hjelm2018learning}. 
% By randomly dividing multiple subjects in a mini-batch, mutual information among the subjects can be estimated through the DIM framework. Further, the maximizing of the mutual information leads to minimizing of the discrepancy among subjects. 
% \todo{Thus, we introduce a method to encourage a feature representation to be subject-invariant by maximizing mutual information among all subjects.} 
% \rebuttal{To maximize the mutual information analogous by using DIM \cite{hjelm2018learning}, we use two types of discriminator $T$, \ie, local and global. Specifically,} we estimate and maximize the mutual information between the class-relevant and global features by using local discriminator $T_\text{l}$ and global discriminator $T_\text{g}$. 
\HI{While the two networks $T_{\text{l}}$ and $T_{\text{g}}$ take the same inputs of $\mathbf{f}_\text{re}$ and $\mathbf{f}_{\text{g}}$, they consider different levels of information in mutual information estimation. The network $T_{\text{l}}$ estimates the mutual information in a fine-grained \emph{local} level for each one of the patches\footnote{\HI{Here, one patch denotes a depth-wise vector in a tensor.}} in the feature $\mathbf{f}_{\text{re}}$ with respect to the global feature $\mathbf{f}_{\text{g}}$ as follows:
\begin{equation}\label{eq:mi_local}
    \max_{E_\text{l},E_\text{g},V,T_\text{l}}\frac{1}{h_1\times w_1}\sum_{i=1}^{N}\sum_{j=1}^{h_1\times w_1}\mathcal{I}_\text{l}(\mathbf{f}_\text{g}^{(i)}\ej{;}\mathbf{f}_\text{re}^{(i,j)})
\end{equation}
where $\mathbf{f}_\text{re}^{(i,j)}$ denotes a $j$-th local patch of \rebuttal{the class-relevant feature $\mathbf{f}_\text{re}^{(i)}$}.
In the meantime, the network $T_{\text{g}}$ estimates in a coarse \emph{global} level by jointly considering all patches in the feature $\mathbf{f}_\text{re}$ as follows:
\begin{equation}\label{eq:mi_global}
    \max_{E_\text{l},E_\text{g},V,T_\text{g}}\sum_{i=1}^{N}\mathcal{I}_\text{g}(\mathbf{f}_\text{g}^{(i)}\ej{;}\mathbf{f}_\text{re}^{(i)})
\end{equation}
} 

After calculating the mutual information between the global feature \HI{$\mathbf{f}_\text{g}^{(i)}$ and each of the local patches $\mathbf{f}_\text{re}^{(i,j)}$ in the class-relevant feature $\mathbf{f}_\text{re}$ in Eq. \eqref{eq:mi_local}, their averaged values is} used to update the parameters of the local encoder $E_\text{l}$, \ejj{the} global encoder $E_\text{g}$, \ejj{the} point-wise convolutional layer $V$, and \ejj{the} $T_\text{l}$. As a result, \ejj{the} global feature $\mathbf{f}_\text{g}$ includes more information with regards to the local regions of the input; thus, it \HI{enriches the representation} in terms of data quality \cite{hjelm2018learning}. \rebuttal{Note that we randomly select trials from a number of subjects and use them for a mini-batch \HI{during training}. Thus, \HI{the networks of $T_{\text{l}}$ and $T_{\text{g}}$ share the same information from} multiple subjects within a mini-batch. As a result,} owing to the maximization of the mutual information between the class-relevant feature and the global feature among multiple subjects in two different ways, the calculated global feature \HI{$\mathbf{f}_\text{g}^{(i)}$ becomes more subject-invariant and contain enriched feature representation for classification}.

\subsection{Objective Function}
\rebuttal{All three objectives can be jointly used to train all components in an end-to-end manner. Our overall objective function $\HI{\mathcal{J}}$ is defined as follows: 
\begin{equation}\label{eq:overall}
    \HI{\mathcal{J}=}\ \alpha\mathcal{L}_\text{cls} + \beta\mathcal{L}_\text{dec} + \gamma\mathcal{L}_\text{DIM} 
\end{equation}
where $\mathcal{L}_\text{DIM}$ is a sum of Eq. \eqref{eq:mi_local} and Eq. \eqref{eq:mi_global}. $\alpha$, $\beta$, and $\gamma$ are hyper-parameters to control the balance \HI{among} three loss terms.}

\section{Experiments \& Results}
We expect that our proposed network can be generalized for multiple subjects by learning subject-invariant and class-relevant representations. In this regard, we validated our methods by considering the following two scenarios (or applications): (i) \rebuttal{a possibility of learning generalizable representation across sessions or subjects} and (ii) a \rebuttal{feasibility of \HI{zero-training} approach} of an unseen subject. We evaluated the proposed method over two public large datasets: \eg, GIST \cite{cho2017eeg} and KU \cite{lee2019eeg} motor imagery datasets. The codes are available at \url{https://github.com/eunjin93/SICR_BCI}.

% We expect that our proposed network can be generalized for multiple subjects by learning subject-invariant and class-relevant representation\edit{s}. In this regard, we validated our methods by considering the following two scenarios (or applications): (i) \rebuttal{} \edit{performance enhancement} due to the effect of data augmentation from multiple patterns of subjects\edit{, and} (ii) a possibility of transfer learning \edit{of} an unseen subject with a small amount of data. We evaluated the proposed method over two public \edit{large} datasets: \eg, GIST- \cite{cho2017eeg} and KU-motor imagery dataset\edit{s} \cite{lee2019eeg}. \revised{\edit{The codes} are \edit{available} at \url{https://github.com/eunjin93/SICR_BCI}.}

\subsection{Data \& Preprocessing}

\subsubsection{GIST dataset \cite{cho2017eeg}}
This\footnote{Available at \url{http://gigadb.org/dataset/100295}} is a big dataset of $52$ subjects and comprises EEG signals related to two different motor imagery tasks of the left- and right-hands. All EEG signals were recorded from $64$ Ag/AgCl electrodes according to the $10$-$20$ system and sampled at $512$ Hz. Each class of a subject comprises $100$ or $120$ trials acquired from four sessions. All subjects were asked to take a rest for $2$ s and then imagine the hand movement for $3$ s by following the instruction given on the monitor. Since two subjects (subjects 29 and 34) had a high correlation with electromyography (EMG), they were termed as bad subjects and their data was not considered in the analysis \cite{cho2017eeg}. Thus, we conducted experiments using the \edit{data of the remaining} $50$ subjects.

\subsubsection{KU dataset \cite{lee2019eeg}}
This\footnote{Available at \url{http://gigadb.org/dataset/10054}}  dataset contains \edit{EEG signals of $54$ subjects for} two motor imagery tasks, \ie, \edit{of the left- and right-hands}, recorded from $62$ Ag/AgCl electrodes according to \edit{the} $10$-$20$ system and sampled \edit{at} $1$k Hz. All EEG signals are acquired from two sessions. Unlike \edit{the} GIST-motor imagery dataset \cite{cho2017eeg}, \edit{the KU} dataset is divided into training and test phase\edit{s} for each session. In each session, each subject \edit{undergoes} $100$ trials per class regardless of the phase. All subjects took a rest \edit{of} $3$ s, and then performed the imagery task of hand movement for $4$ s by following the given instruction in the monitor. \rebuttal{For the low computational cost, we downsampled EEG signals to 500 Hz.}

\subsubsection{Preprocessing}
\revised{For} both datasets, the signals were preprocessed \edit{using} a large Laplacian \edit{filter} to reduce noise and a bandpass \edit{filter} in the range of $4-40$ Hz related to sensorimotor rhythms. After segmenting the signals to a baseline and task, we discarded \edit{the} first and last $0.5$ second\edit{s of} the segmented task signal. \edit{Next}, we subtracted \edit{the} mean of the baseline to each task signal for baseline correction. 
% In an attempt to exploit frequency properties of the data, we computed the PSD by applying Welch's method. Consequently, we obtained \revised{[$(64\text{ or }62)\times37$]}-sized data per sample and used \edit{them} as an input of our method. 

\subsection{Experimental Settings}
\rebuttal{Although each subject in both GIST \cite{cho2017eeg} and KU \cite{lee2019eeg} datasets performed motor imagery tasks over different sessions, we simply combined samples across sessions into one. To verify the generalization of the network, we conducted experiments considering the following two scenarios:} 
\begin{itemize}
    \item \rebuttal{Scenario I (cross-subject learning): In this scenario, we considered the use of all available samples from other subjects as well as a target subject. The underlying assumption behind this scenario is that it could be possible to learn better feature representations based on diverse EEG signals.} 

    \rebuttal{In order for performance evaluation, we divided the samples of each subject into 5 folds by assigning an equal number of samples to each fold, and used samples of 4 folds for training and samples of one remaining fold for test. The training samples of 4 folds were further randomly partitioned into a pure training set and an independent validation set at a ratio of $7:1$. We trained and selected models with combined training and validation samples of all subjects, respectively, and then tested on the testing samples \HI{of} each subject. That is, a single subject-independent model was built and tested on all subjects. The processes were repeated 5 times and their average performance was reported for evaluation.}
    
    % \rebuttal{First, to perform a $5$-fold cross-validation, we randomly divided all trials of each subject into training, validation, and test sets at a ratio of $7:1:2$, respectively.} As our network is considered to learn subject-invariant and class-relevant representations, we trained it using training data of all subjects within a single training session and then tested it \rebuttal{for the test data of each subject}.
    
    \item \rebuttal{Scenario II (zero-training): This is designed for a zero-training application, \ie, no sample of a target subject is used to tune model parameters. That is, we first trained a model based on solely samples \HI{of non-target} subjects, and then tested the trained model on the samples of a new (unseen) target subject. For this, we have exploited a leave-one-subject-out \HI{validation} scheme.}
    
    % \rebuttal{After training our proposed network, it could develop common features among subjects. Thus, we can take advantage of this characteristic for a new subject. Without a target subject, we trained our network by using all training samples of the remaining subjects and then tested it using the whole samples of the new target subject. In other words, we used leave-one-subject-out cross-validation strategy in Scenario II.}
\end{itemize}

\subsection{Model Implementation}
\subsubsection{Architecture}
\rebuttal{To demonstrate that our method can be independent of the architecture of the feature extractor, we utilized two deep learning-based motor imagery decoding models as our encoder.}
\begin{itemize}
    \item \rebuttal{Deep ConvNet \cite{schirrmeister2017deep}: This is composed of four convolution-pool blocks capturing spatio-temporal information of raw EEG signals and a fully-connected layer. In the first block, an input EEG is convolved with kernels of [$1\times10$] temporally and then convolved with kernels of [$n_c\times1$] to integrate spatial information. The remaining three blocks consist of temporal convolution with [$1\times10$] kernels and max-pooling as [$1\times3$], respectively. For fair comparison, we did not use the ``cropped-training'' strategy which was considered in the original work \cite{schirrmeister2017deep}.}
    \item \rebuttal{EEGNet \cite{lawhern2018eegnet}: This \HI{exploits} depthwise and separable convolutions \cite{chollet2017xception} to encode EEG signals for reduction of the number of parameters. In detail, EEGNet \cite{lawhern2018eegnet} consists of three convolutional layers: (i) the first convolutional layer has filters of [$1\times f_s/2$], where $f_s$ denotes the sampling rate of the data and then encodes the temporal information of the inputs, (ii) the second convolutional layer convolves kernels of [$n_c\times1$] to the output of the first layer in a depth-wise manner \cite{chollet2017xception}, and (iii) the separable convolution \cite{chollet2017xception} is utilized as the last layer to summarize the temporal information of each feature map.}
    % \item \rebuttal{RSTNN?}
    % \item \rebuttal{MSNN?}
\end{itemize}
\rebuttal{Since both Deep ConvNet \cite{schirrmeister2017deep} and EEGNet \cite{lawhern2018eegnet} include several convolutional layers, we set the last layer as our global encoder and the previous layers as our local encoder.} Note that we modified the spatial filter size as the number of electrode channels in each dataset. \rebuttal{After obtaining the local feature, we conducted a point-wise convolution to expand the depth dimension double. We split the embedded local feature along the depth-axis to divide it into the class-relevant feature and the class-irrelevant feature.}

\rebuttal{In the $T_\text{g}$, after embedding the class-relevant feature through another convolutional layer which has same architecture with the global encoder, the feature was concatenated with the global feature along the depth-axis. Then, the concatenated feature was taken as the input of a fully-connected layer that outputs $1$ unit. We used the \emph{concat-and-convolve architecture} for the $T_\text{l}$ in accordance with \cite{hjelm2018learning}. In the $T_\text{l}$, the global feature was replicated to be the same dimension with the class-relevant feature and then concatenated with the class-relevant feature at every location. Subsequently, the concatenated feature were passed to the pointwise convolutional layer and then became same height and width of the class-relevant feature. After attaining scores corresponding the joint distribution and the product of marginals between the class-relevant feature and the global feature, we calculated their average to use in Eq. \eqref{eq:mi_local}. The classifier $C$ was composed of a fully-connected layer with the number of classes units. More details can be found in Supplementary A.}

\subsubsection{Training Settings} Exponential linear units (ELU) were used as a nonlinear function in our network. In addition to the two networks, \ie, $T_\text{g}$ and $T_\text{l}$, we applied a batch normalization \cite{ioffe2015batch}. \rebuttal{Furthermore, we applied an $l_2$-regularization with a coefficient of \rebuttal{$0.1$} and a dropout \cite{srivastava2014dropout} with a rate of $0.5$ to prevent over-fitting. We trained models by using a RAdam \cite{liu2019radam} with a learning rate of $10^{-3}$ by exponentially decreasing $0.99$ per epoch, where the dimension of a mini-batch size was $40$. Regarding the hyper-parameters $\alpha$, $\beta$ and $\gamma$ in Eq. \eqref{eq:overall}, we chose $\{\alpha, \beta, \gamma\}$ as $\{0.5,0.3,0.5\}$ for all cases.}

\subsection{Competing Methods}
\rebuttal{For evaluation, we compared our proposed method with the following various methods using the same architecture of the feature extractor:
\begin{itemize}
    \item Pooled learning: \rebuttal{In order to see the effectiveness of a transfer learning approach, we first considered the most straightforward way of using all available samples \HI{from many subjects} by pooling them into a single large dataset, as a baseline. In this case, we did not explicitly model subject-invariance or class-relevance of features, but took advantage of all available samples in feature representation and classifier learning. We implemented conventional machine learning-based models (\eg, CSP \cite{ramoser2000optimal} and FBCSP \cite{ang2008filter}) as well as deep learning-based models.} 
    \item Subject-to-Subject transfer learning: Various forms of regularized CSPs (RCSPs) \cite{lotte2010regularizing}, which exploit covariance matrices derived from other subjects' EEG signals for better generalization in a way of regularization, have been studied and presented reasonable performance in subject-to-subject transfer \cite{kang2009composite,lotte2010regularizing}. Among those, we implemented Lotte \etal's Weighted Tikhonov Regularization method (WTRCSP) because of its superiority to its counterpart methods in performance \cite{lotte2010regularizing}. Subsequently, we used LDA based \HI{classifier} by taking the WTRCSP features as input.% \cmt{\cite{vidaurre2010toward}}.% and then tested it for the target subject's test samples.
    \item Domain adversarial neural network for motor imagery signals (DANN-MI) \cite{ozdenizci2020learning}: Inspired by DANN \cite{ganin2016domain}, \"{O}zdenizci \etal~\cite{ozdenizci2020learning} introduced a deep EEG feature \HI{representation learning robust} across intra- and inter-subject variability via adversarial inference. DANN-MI composed of three modules, \ie, a feature extractor, a classifier, and an adversary, and trained by two loss terms, \ie, a classification loss and a subject discrimination loss. Specifically, based on the adversarial learning between the adversary and the feature extractor, their feature extractor learned class-discriminative and subject-invariant feature representation. \cite{ozdenizci2020learning} also set the hyper-parameter to tune the importance between two losses. We chose $\lambda$ as $0.03$ for EEGNet \cite{lawhern2018eegnet} and $0.05$ for Deep ConvNet \cite{schirrmeister2017deep} according to the original work \cite{ozdenizci2020learning}.
    \item Deep adversarial disentangled autoencoder (DADA) \cite{peng2019domain}: \HI{Because of the philosophical similarity to our method, we also compared with DADA, which was} also inspired by DANN \cite{ganin2016domain} and devised to tackle a domain adaptation issue in computer vision. DADA disentangles the domain-invariant features from both domain-specific and class-irrelevant features simultaneously to prevent negative transfer \cite{peng2019domain}. 
\end{itemize}}

\begin{table}[t]
    \centering
     \caption{\rebuttal{Averaged performance [$\%$] under Scenario I.} ($*$: $p<0.05$, $\dagger$: no statistical difference)}
     \begin{center}\footnotesize
         (a) \HI{Comparison with linear methods in pooled or transfer learning and deep-learning methods in pooled learning.}
     \end{center}
\rebuttal{\begin{tabular}{ccc}\toprule
 & GIST dataset \cite{cho2017eeg} & KU dataset \cite{lee2019eeg} \\\toprule
CSP \cite{ramoser2000optimal} (Pooled learning)                  &   $41.53\pm8.40$\textsuperscript{$*$}                       &    $56.83\pm7.72$\textsuperscript{$*$}                    \\
FBCSP \cite{ang2008filter} (Pooled learning)               &  $46.24\pm6.84$\textsuperscript{$*$}                        &       $44.67\pm7.10$\textsuperscript{$*$}                 \\
WTRCSP \cite{lotte2010regularizing} \HI{(Transfer learning)}             & $55.84\pm10.25$\textsuperscript{$*$}                         &        $55.10\pm8.98$\textsuperscript{$*$}                \\
Deep ConvNet \cite{schirrmeister2017deep} (Pooled learning) & $54.77\pm8.54$\textsuperscript{$*$} & $58.45\pm11.64$\textsuperscript{$*$}  \\
EEGNet \cite{lawhern2018eegnet} (Pooled learning) & $55.38\pm9.55$\textsuperscript{$*$} & $64.67\pm13.24$\textsuperscript{$*$} \\
Ours (\HI{with} Deep ConvNet \cite{schirrmeister2017deep})    & $74.15\pm12.64$  &    $\mathbf{76.67\pm13.01}$                 \\
Ours (\HI{with} EEGNet \cite{lawhern2018eegnet}) & $\mathbf{76.60\pm12.48}$ & $74.48\pm13.84$ \\\bottomrule  
\end{tabular}}
     \begin{center}\footnotesize
\rebuttal{        (b) \HI{Comparison with methods of} deep learning-based\\ transfer learning on GIST dataset \cite{cho2017eeg}.}
     \end{center}
\rebuttal{\begin{tabular}{ccc}\toprule
    & Deep ConvNet \cite{schirrmeister2017deep}    & EEGNet \cite{lawhern2018eegnet}     \\\toprule
% \cmt{Pooled learning} & $54.77\pm8.54$\textsuperscript{$*$} & $55.38\pm9.55$\textsuperscript{$*$} \\
% \ejj{WTRCSP \cite{lotte2010regularizing}} &  \multicolumn{2}{c}{$55.84\pm10.25$\textsuperscript{$*$}} \\
DANN-MI \cite{ozdenizci2020learning}  & $73.17\pm12.81$\textsuperscript{$\dagger$}            &  $74.92\pm12.44$\textsuperscript{$*$}                 \\
DADA [4]     &    $50.26\pm2.76$\textsuperscript{$*$}          & $72.12\pm13.31$\textsuperscript{$*$}                 \\ 
Ours                                &  $\mathbf{74.15\pm12.64}$            & $\mathbf{76.60\pm12.48}$              \\ \bottomrule   
\end{tabular}}

\begin{center}\footnotesize
\rebuttal{(c) \HI{Comparison with methods of} deep learning-based\\ transfer learning on KU dataset \cite{lee2019eeg}.}
\end{center}
\rebuttal{\begin{tabular}{ccc}
    \toprule
                                    & Deep ConvNet \cite{schirrmeister2017deep}       & EEGNet \cite{lawhern2018eegnet}      \\\toprule
% \cmt{Pooled learning} & $58.45\pm11.64$\textsuperscript{$*$} & $64.67\pm13.24$\textsuperscript{$*$} \\
% \ejj{WTRCSP \cite{lotte2010regularizing}} & \multicolumn{2}{c}{$55.10\pm8.98$\textsuperscript{$*$}}\\
DANN-MI \cite{ozdenizci2020learning}  & $74.96\pm13.15$\textsuperscript{$*$}              &  $72.43\pm13.90$\textsuperscript{$*$}                  \\
DADA \cite{peng2019domain}    & $50.93\pm2.66$\textsuperscript{$*$}              &     $68.43\pm13.75$\textsuperscript{$*$}             \\ 
Ours                                & $\mathbf{76.67\pm13.01}$             &   $\mathbf{74.48\pm13.84}$           \\ \bottomrule   
\end{tabular}}
    \label{tab:scenario1}
\end{table}

\begin{table}[t]
    \centering
     \caption{\rebuttal{Averaged performance [$\%$] under Scenario II.} ($*$: $p<0.05$, $\dagger$: no statistical difference)}
     \begin{center}\footnotesize
\rebuttal{        (a) GIST dataset \cite{cho2017eeg}   }
     \end{center}
\rebuttal{\begin{tabular}{ccc}
    \toprule
                                    & Deep ConvNet \cite{schirrmeister2017deep}       & EEGNet \cite{lawhern2018eegnet}        \\\toprule
DANN-MI \cite{ozdenizci2020learning}  & $69.64\pm12.93$\textsuperscript{$\dagger$}               &    $72.56\pm13.61$\textsuperscript{$\dagger$} \\
% DADA \cite{peng2019domain}                                &              &                  \\ 
Ours                                & $\mathbf{70.54\pm12.84}$  &    $\mathbf{73.73\pm13.75}$  \\ \bottomrule   
\end{tabular}}
\begin{center}\footnotesize
\rebuttal{(b) KU dataset \cite{lee2019eeg}}
\end{center}
\rebuttal{    \begin{tabular}{ccc}
    \toprule
                                    & Deep ConvNet \cite{schirrmeister2017deep}       & EEGNet \cite{lawhern2018eegnet}       \\\toprule
DANN-MI \cite{ozdenizci2020learning}     & $69.41\pm13.17$\textsuperscript{$*$}             &  $72.08\pm14.29$\textsuperscript{$*$}                  \\
% DADA \cite{peng2019domain}                                &              &                  \\ 
Ours                                &   $\mathbf{73.32\pm13.55}$           &     $\mathbf{72.22\pm13.51}$           \\ \bottomrule   
\end{tabular}}
% \begin{tablenotes}
% \scriptsize
%     \item \rebuttal{$*$: $p<0.05$, $\dagger$: no statistical difference}
% \end{tablenotes}
    \label{tab:scenario2}
\end{table}

\begin{table}[t]    
\centering
     \caption{\HI{Performance comparison in an ablation study of different components in our proposed framework under} Scenario I. ($*$: $p<0.05$, $\dagger$: no statistical difference)}
     \begin{center}\footnotesize
        (a) GIST dataset \cite{cho2017eeg}          
     \end{center}
    \rebuttal{\begin{tabular}{ccc}
    \toprule
                                    & Deep ConvNet \cite{schirrmeister2017deep}       & EEGNet \cite{lawhern2018eegnet}        \\\toprule
Model \textbf{I}   &   $73.31\pm12.68$\textsuperscript{$\dagger$}           &  $74.48\pm12.87$\textsuperscript{$*$}           \\
Model \textbf{II}  &    $73.68\pm12.05$\textsuperscript{$\dagger$}          &  $75.94\pm13.20$\textsuperscript{$\dagger$}             \\
Model \textbf{III} &    $73.34\pm12.44$\textsuperscript{$\dagger$}          & $75.44\pm12.93$\textsuperscript{$*$}            \\
Model \textbf{IV}     &  $\mathbf{74.15\pm12.64}$            & $\mathbf{76.60\pm12.48}$              \\ \bottomrule   
\end{tabular}}

\begin{center}\footnotesize
(b) KU dataset \cite{lee2019eeg}
\end{center}
    \rebuttal{\begin{tabular}{ccc}
    \toprule
                                    & Deep ConvNet \cite{schirrmeister2017deep}       & EEGNet \cite{lawhern2018eegnet}        \\\toprule
Model \textbf{I}   &    $75.22\pm13.42$\textsuperscript{$*$}        &    $73.26\pm14.12$\textsuperscript{$*$}         \\
Model \textbf{II}  &    $75.70\pm13.30$\textsuperscript{$*$}          &  $74.06\pm13.49$\textsuperscript{$\dagger$}            \\
Model \textbf{III} &    $75.45\pm13.20$\textsuperscript{$*$}          &  $73.41\pm13.85$\textsuperscript{$*$}           \\
Model \textbf{IV} & $\mathbf{76.67\pm13.01}$             &   $\mathbf{74.48\pm13.84}$           \\ \bottomrule   
\end{tabular}}
% \begin{tablenotes}
% \scriptsize
%     \item $*$: $p<0.05$, $\dagger$: no statistical difference
%     % \item \textsuperscript{$*$}: $p<0.05$
% \end{tablenotes}
    \label{tab:Ablation}
\end{table}

\subsection{Ablation study} 
\rebuttal{In addition, to validate the effectiveness of each component in our model, we performed three different ablations under Scenario I. First, we trained our proposed method by considering only the class decomposition step, referred as Model \textbf{I}. In other words, Model \textbf{I} was trained by using the \HI{composite of a classification loss in Eq. \eqref{eq:l_cls} and a decomposition loss in Eq. \eqref{eq:mine}}. We developed Model \textbf{II} composed of encoders ($E_\text{l},E_\text{g}$ and $V$), $T_\text{g}$, and $C$ and trained it by \HI{jointly} minimizing the classification loss in Eq. \eqref{eq:l_cls}, the decomposition loss in Eq. \eqref{eq:mine} and the global MINE loss in Eq. \eqref{eq:mi_global}. Regarding Model \textbf{III}, we removed the $T_\text{g}$ in our \HI{complete network, denoted as as Model \textbf{IV}}. \HI{Concisely,} Model \textbf{III} was trained without the global MINE loss in Eq. \eqref{eq:overall}.}

\begin{figure*}[htp!]
    \centering
    \begin{center}\footnotesize\label{fig:GIST_PSD_LRP_Deep}
    \includegraphics[width=.48\textwidth]{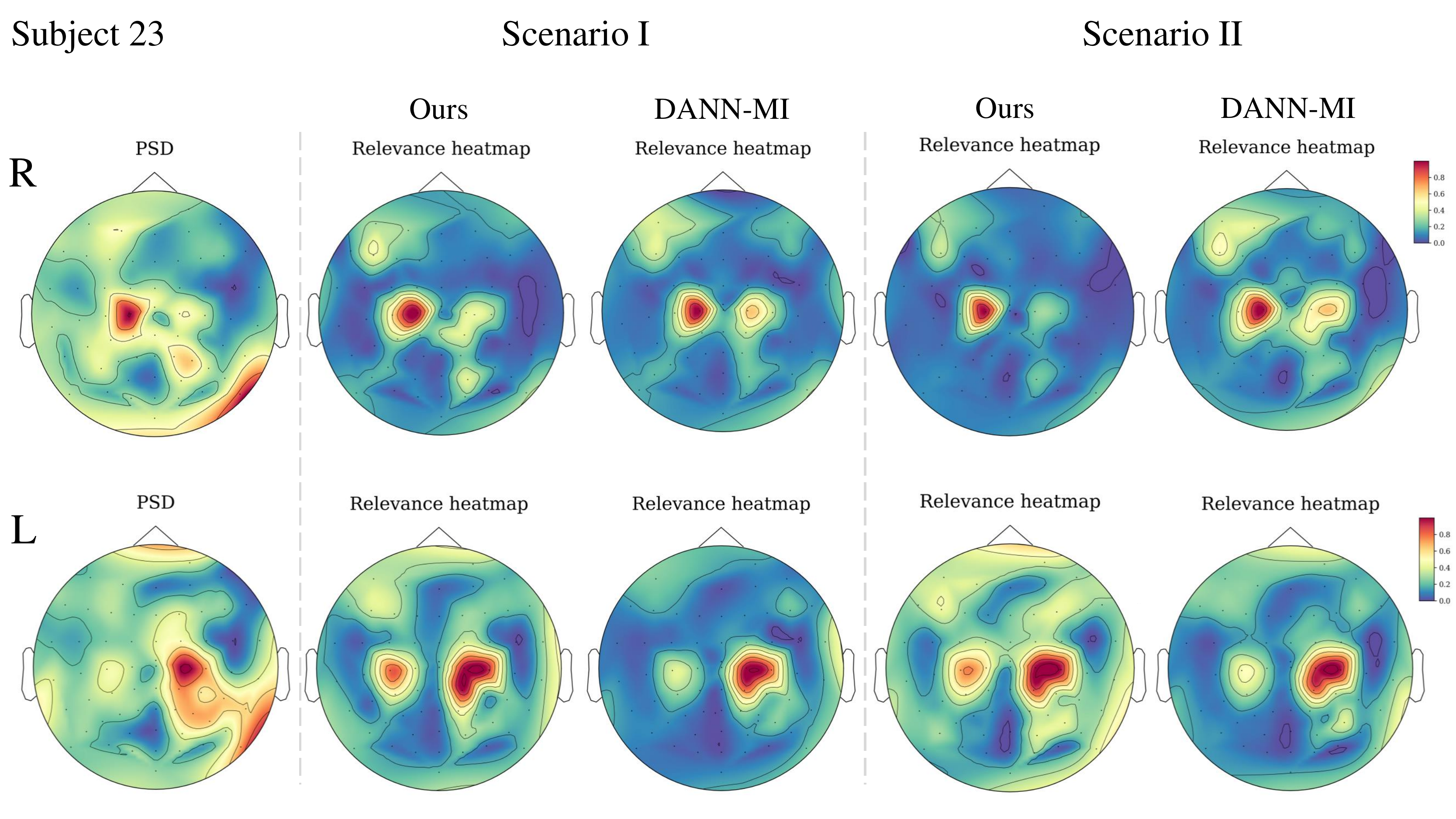}\quad\quad
    \includegraphics[width=.48\textwidth]{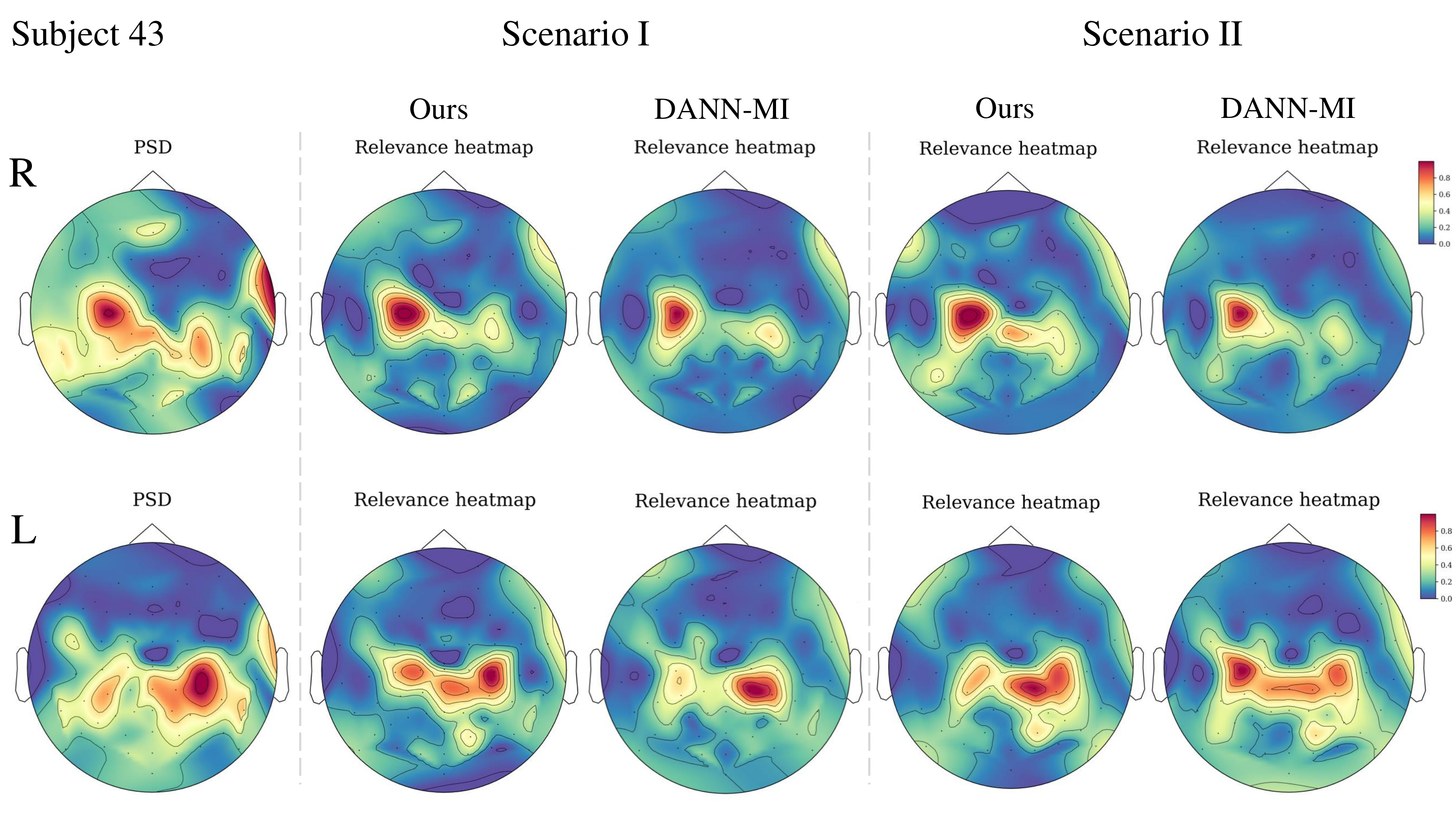}\\
    (a) Deep ConvNet \cite{schirrmeister2017deep} trained on GIST dataset \cite{cho2017eeg}   \quad\quad\quad\quad\quad\quad\quad\quad\quad\quad
    (b) EEGNet \cite{lawhern2018eegnet} trained on GIST dataset \cite{cho2017eeg}\end{center}
    % \begin{center}\footnotesize\label{fig:GIST_PSD_LRP_EEG}
    % \includegraphics[width=.45\textwidth]{TNNLS2019/figures/Final_PSD_LRP_1_GIST_EEGNet_Sub43.pdf}\\
    % (b) EEGNet \cite{lawhern2018eegnet} trained on GIST dataset \cite{cho2017eeg}
    % \end{center}\\
    \begin{center}\footnotesize\label{fig:KU_PSD_LRP_Deep}
    \includegraphics[width=.48\textwidth]{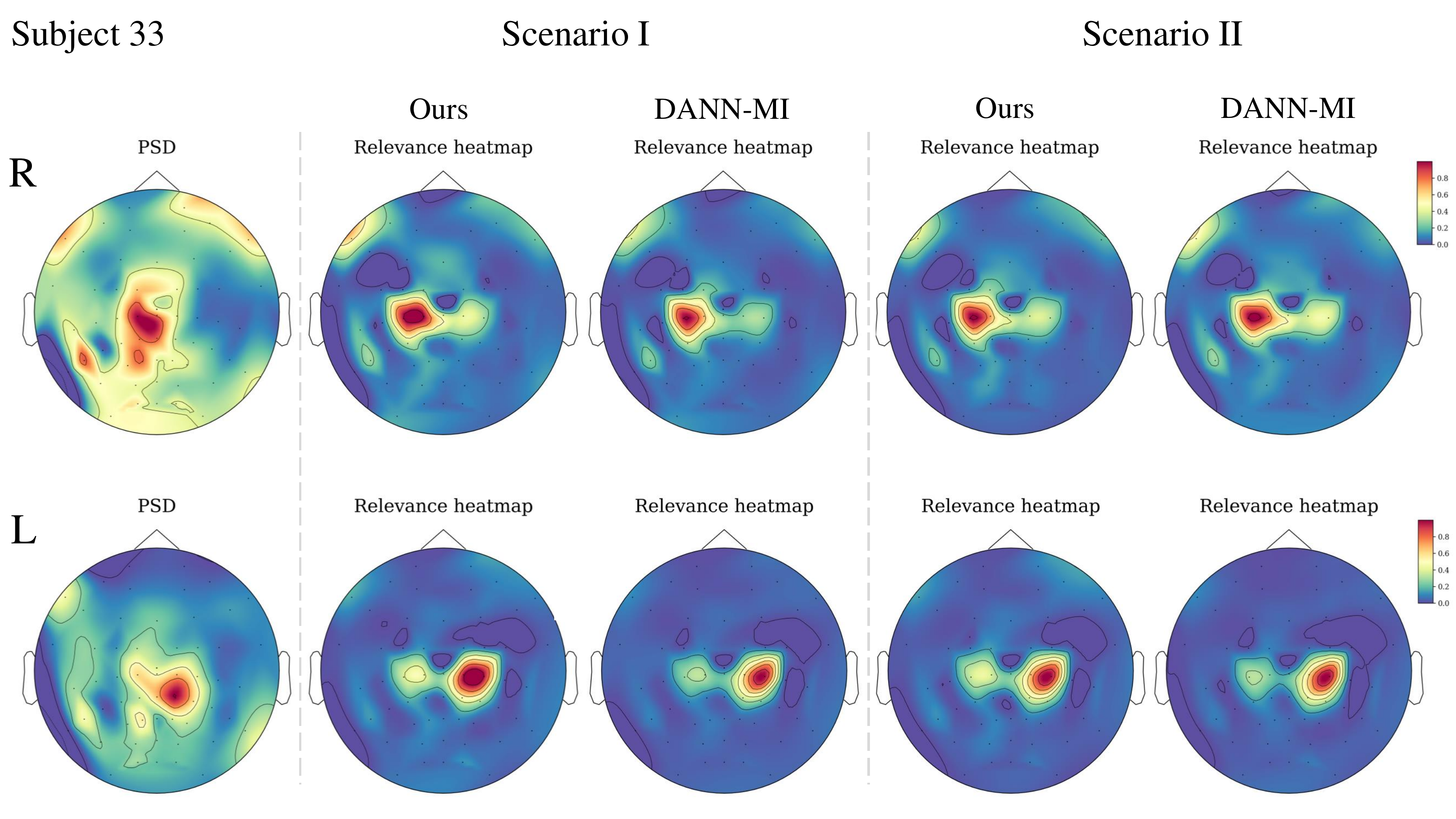}\quad\quad
    \includegraphics[width=.48\textwidth]{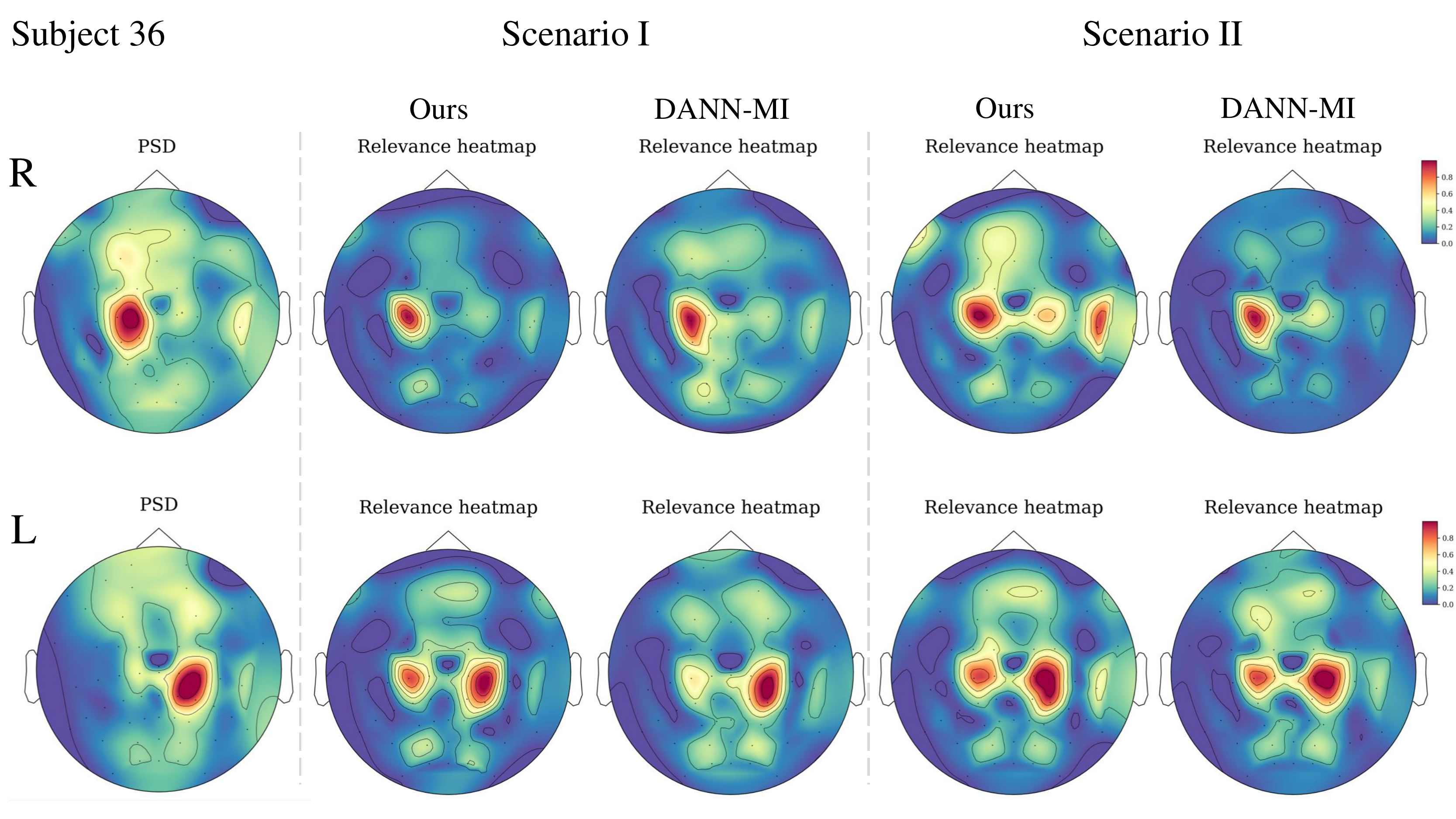}\\
    (c) Deep ConvNet \cite{schirrmeister2017deep} trained on KU dataset \cite{lee2019eeg}
    \quad\quad\quad\quad\quad\quad\quad\quad\quad\quad
    (d) EEGNet \cite{lawhern2018eegnet} trained on KU dataset \cite{lee2019eeg}
    \end{center}
    %     \begin{center}\footnotesize\label{fig:KU_PSD_LRP_EEG}
    % \includegraphics[width=.45\textwidth]{TNNLS2019/figures/Final_PSD_LRP_1_KU_EEGNet_Sub36.pdf}\\
    % (d) EEGNet \cite{lawhern2018eegnet} trained on KU dataset \cite{lee2019eeg}
    % \end{center}
    \caption{\HI{Illustrative comparison of PSD maps and decision-relevance heatmaps \cite{montavon2017explaining} of input EEG signals from randomly selected subjects. We normalized each visualized topoplot in a range between $0$ and $1$. (L: left-hand, R: right-hand)}}
    \label{fig:psd_lrp}
\end{figure*}

\subsection{Results}
\subsubsection{Scenario I} 
\rebuttal{We trained the \HI{comparative methods} with all the subjects of each dataset in \HI{a} subject-independent \HI{way, and tested them with the test samples of each subject}. The averaged performance for all the subjects in two datasets is summarized in TABLE I. Our model showed the accuracy of $74.15\%$ (\HI{with }Deep ConvNet \cite{schirrmeister2017deep}) and $76.60\%$ (\HI{with }EEGNet \cite{lawhern2018eegnet}) on the GIST dataset \cite{cho2017eeg} and achieved the averaged accuracy of $76.67\%$ (\HI{with }Deep ConvNet \cite{schirrmeister2017deep}) and $74.48\%$ (\HI{with }EEGNet \cite{lawhern2018eegnet}) on the KU dataset \cite{lee2019eeg}. Notably, our method outperformed all the competing methods of the pooled learning, DANN-MI \cite{ozdenizci2020learning}, and DADA \cite{peng2019domain} with high statistical significance ($p$-value$<$0.05), except the case of DANN-MI with Deep ConvNet on the GIST dataset \cite{cho2017eeg}.} 
% We trained the network \edit{with} all \edit{the} subjects of each group. The averaged performances of all the subjects in two datasets is summarized in TABLE \ref{tab:scenario1}. \revised{We also plotted the performances of individual subjects among deep learning-based baseline methods\footnote{When comparing individual performance, our results are higher than \edit{those of} the CSP \cite{ramoser2000optimal} and FBCSP \cite{ang2008filter} \edit{methods}.}}. Our model showed $\mathbf{71.7\%}$ and $\mathbf{73.2\%}$ \edit{accuracies} \edit{for} the GIST \cite{cho2017eeg} and KU \cite{lee2019eeg} datasets, on average, over all subjects in each dataset, \edit{thus} outperforming the other motor imagery decoding baseline methods. Especially, from Fig. \ref{fig:result1}, many subjects who are empirically considered as BCI-illiterate subjects \revised{(gray shaded region)} not making sufficient discriminative motor imagery patterns show promising performance improvements \edit{owing} to \edit{model} training \edit{using data of} other subjects including \edit{their} discriminative patterns.

\subsubsection{Scenario II} 
\rebuttal{TABLE II summarizes the results of DANN-MI \cite{ozdenizci2020learning} and our method in Scenario II. Again, our proposed method showed its superiority to DANN-MI \cite{ozdenizci2020learning} in all cases by \HI{moderate} margin, depicting its generalization power to samples of \HI{new (unseen) subjects}. In other words, this supports our hypothesis that the feature representation learning in our framework is effective to be applicable for new \HI{subjects in a way of zero training, \ie, no subject-dependent adaptation or calibration is required}.}

\subsubsection{Ablation Study}
\HI{Table III summarizes the performance differences according to the changes of involving different components in our proposed framework. Although the performance improvement with each individual component was moderate, the joint work of all components showed best performance overall. It is noteworthy that the performance improvements with our proposed full network (Model IV) was statistically significant compared to the counterpart models in several cases. Based on those results, we believe that our feature decomposition and enriched representation learning simultaneously encourage to enhance the performance. For individual performances, refer to  Supplementary B.} 

% \ejj{In TABLE III, our method (\ie, Model \textbf{IV}) showed just a little better than few ablation cases on EEGNet trained GIST dataset \cite{cho2017eeg}. In the meantime, regarding KU dataset \cite{lee2019eeg}, our method was superior to most ablation cases with statistical significance. However, the joint work of all components showed the best performance with statistical significance in some cases. Thus, we believe that our feature decomposition and enriched representation learning simultaneously encourage to enhance the performance.}

% \begin{figure}[t]
%     \centering
%     \begin{center}\footnotesize\label{fig:GIST_PSD_LRP}
%     \includegraphics[width=.5\textwidth]{TNNLS2019/figures/GIST_PSD_LRP.pdf}
%     (a) GIST dataset \cite{cho2017eeg}
%     \end{center}
%     \begin{center}\footnotesize\label{fig:KU_PSD_LRP}
%     \includegraphics[width=.5\textwidth]{TNNLS2019/figures/KU_PSD_LRP.pdf}        
%     (b) KU dataset \cite{lee2019eeg}
%     \end{center}
%     \caption{Illustration of PSD and LRP heatmaps \cite{montavon2017explaining}. \rebuttal{In each figure, first row presents the right-hand class and second row denotes the left-hand class. We normalized each visualized topoplot in a range between $0$ and $1$.}}
%     \label{fig:psd_lrp}
% \end{figure}

\section{Analyses \& Discussion}
This section presents the \ejj{analyses} of our proposed network from two aspects: neurophysiological explanation and feature decomposition. First, we performed layer-wise relevance propagation (LRP) \cite{montavon2017explaining} to obtain an insight into the neurophysiological explanation. Next, from the viewpoint of feature decomposition, we plotted the learned feature by using t-distributed stochastic neighbor embedding (t-SNE) \cite{maaten2008visualizing} to determine how the class-relevant and class-irrelevant features are distributed. 
% In addition to t-SNE \cite{maaten2008visualizing}, we also conducted ablation studies to show the validity of each component in our network.

\subsection{Neurophysiological Explanation}
\rebuttal{LRP \cite{montavon2017explaining} has been \HI{used} in several studies in BCI to interpret their results neurophysiologically \cite{ozdenizci2020learning}.  We also applied LRP to explain which channels of an input EEG contribute to the decision in our proposed network. We implemented LRPs by following $\epsilon$-rule to calculate relevance scores \cite{bach2015pixel} and then compared them to the PSD of raw motor imagery EEGs by using each subject's test samples for both Scenario I and Scenario II. First, we computed the PSD by conducting Welch's method \cite{welch1967use}. In an attempt to exploit spatial properties of EEG, we averaged the PSD and relevance scores in the frequency and time axis, respectively. Fig. 3 plots PSD maps and relevance heatmaps \HI{of randomly selected subjects} for each class in a topological manner. \HI{Notably, in both PSD maps and relevance heatmaps,} it is clearly observable the contra-lateral patterns. That is, when imagining left-/right-hand movements, the right/left motor-related areas \HI{showed noticeable activation patterns} (red color), respectively. \HI{Those activated areas were highly contributed to making a decision by our method, as illustrated in the respective relevance heatmaps}. Although our proposed method removed partial information considered as the class-irrelevant representation, both feature extractors (EEGNet \cite{lawhern2018eegnet} and Deep ConvNet \cite{schirrmeister2017deep}) could still learn a spatio-spectral pattern inherent in motor imagery EEGs. With regard to the relevance heatmaps obtained in Scenario II, even though our network did not leverage the target subject's samples, it could detect patterns from motor-related areas that served to identify the class of input EEGs. As a result, we believe our model well learned and utilized the neurophysiological characteristics related to motor imagery, by learning subject-invariant and class-relevant feature representations.}

\begin{figure}[t]
    \centering
    \begin{center}\footnotesize\label{fig:GIST_Deep_t-SNE}
    \includegraphics[width=.46\textwidth]{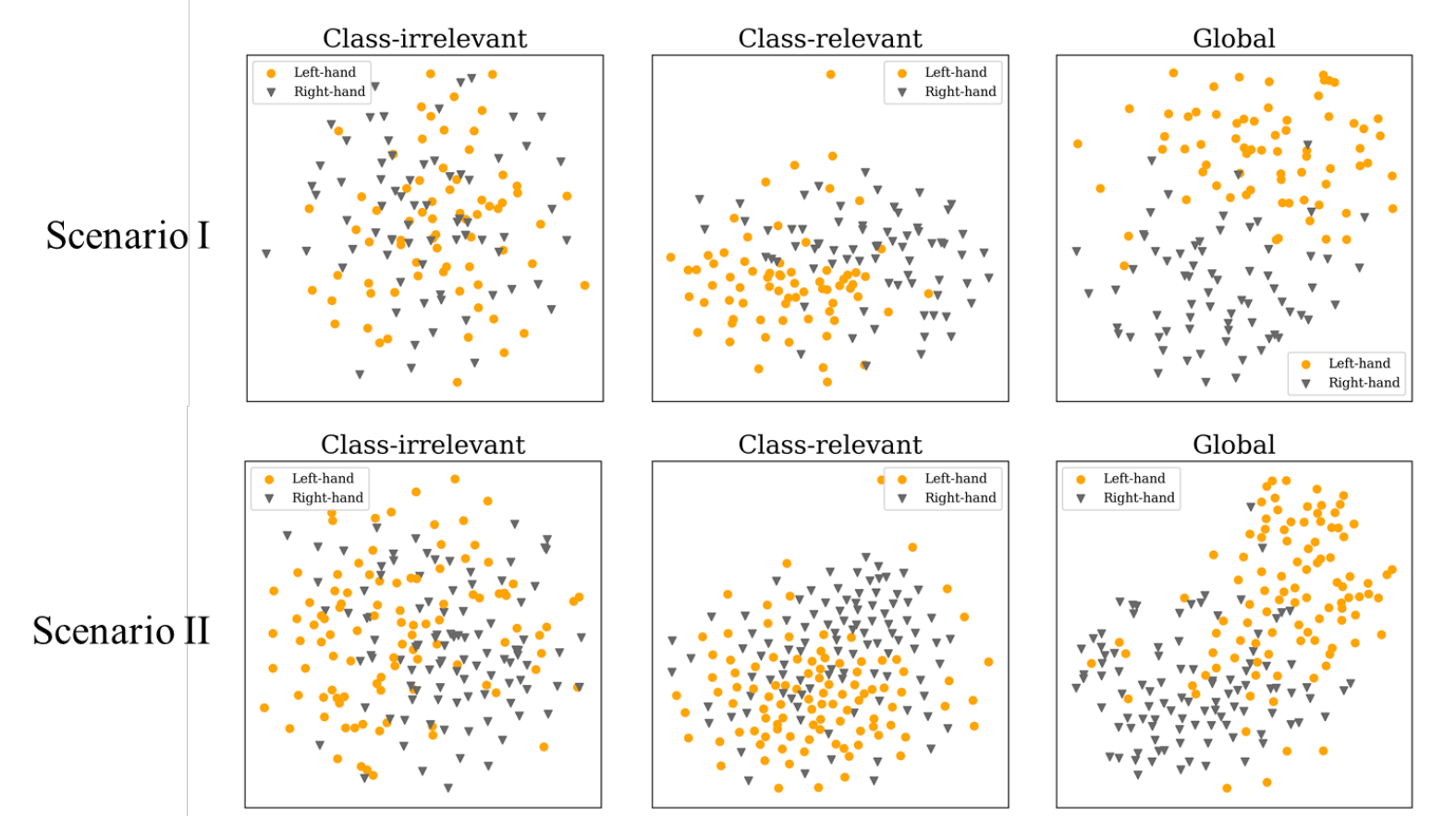}\\
    (a) Subject $5$ of GIST dataset \cite{cho2017eeg}  
    \end{center}
    \begin{center}\footnotesize\label{fig:KU_EEGNet_t-SNE}
    \includegraphics[width=.46\textwidth]{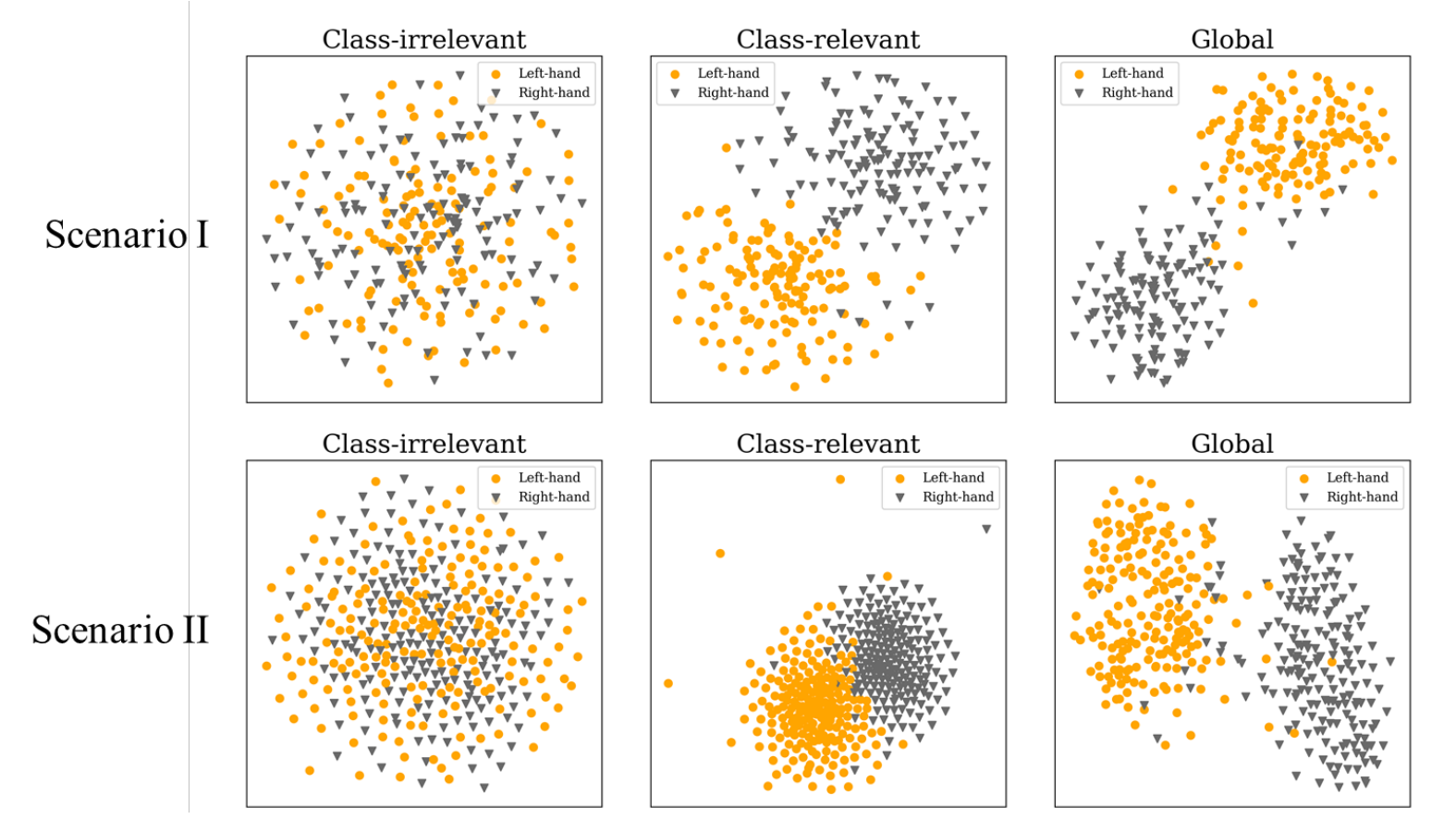}\\
    (b) Subject $21$ of KU dataset \cite{lee2019eeg}
    \end{center}
    \caption{\rebuttal{t-SNE \cite{maaten2008visualizing} of the class-irrelevant, class-relevant, and global features in Scenario I and II. We visualized all features trained using Deep ConvNet \cite{schirrmeister2017deep}.}}
    \label{fig:t-SNE}
\end{figure}

\subsection{Feature Decomposition}
\rebuttal{We used t-SNE \cite{maaten2008visualizing} to visualize the class-irrelevant, class-relevant, and global features of subject $5$ of GIST dataset \cite{cho2017eeg} and subject $21$ of KU dataset \cite{lee2019eeg} under both Scenario I and Scenario II as shown in Fig. \ref{fig:t-SNE}. We observed that compared with the class-irrelevant features, the class-relevant features and the global features were separable between two classes, \ie, for the left- and right-hands. Note that in comparison between distributions of class-relevant and global features, that of global features enriched by means of local-global mutual-information maximization was better separable showing relatively large between-class distance. In addition, although we did not exploit a target subject's data for training, the target subject's samples were also highly distinguishable between two classes in Scenario II. Hence, we argue that our proposed method could find class-relevant and subject-invariant features, and successfully used to classify samples of unseen subjects.}

% \begin{figure}[t]
%     \centering
%     \includegraphics[scale=0.32]{TNNLS2019/figures/TNNLS_Figure2_t-SNE_GIST2_S1.pdf}
% \caption{t-SNE \cite{maaten2008visualizing} of the class-relevant and the class-irrelevant features in Scenario I.}
% \label{fig:tsne_all}
% \end{figure}

% \begin{table}[h]
%     \centering
%     \caption{Data \rebuttal{accuracy} of all subjects in of each dataset in Scenario I. \textsuperscript{*} denotes $p<0.001$.}
%     \begin{tabularx}{0.7\linewidth}{ccc}
% \toprule
%                     & GIST dataset \cite{cho2017eeg}& KU dataset \cite{lee2019eeg} \\ \toprule
% Model \textbf{I}\textsuperscript{*}            & $65.4\pm10.2$     & $69.3\pm13.1$   \\ 
% Model \textbf{II}\textsuperscript{*} & $65.3\pm10.4$     &   $69.2\pm13.0$  \\ 
% % Model \textbf{III}\textsuperscript{*}  & $68.6\pm11.1$      &  $70.0\pm12.7$  \\ \midrule
% \textbf{Model} \textbf{III} & $\mathbf{71.7\pm9.2}$     &    $\mathbf{73.2\pm11.9}$\\ \bottomrule
% \end{tabularx}
%     \label{tab:Ablation}
% \end{table}

\section{Conclusion}
In this work, we proposed a deep neural network that learns subject-invariant and class-relevant representations via mutual information estimation among features in different levels for BCI tasks in an end-to-end manner. \rebuttal{The subject-invariant and class-relevant feature space can be deployed for decoding a new subject and mitigate negative transfer. Notably, we showed our proposed method is independent of the feature representation, thus, we can utilize any existing decoding models as our feature extractor.}
% The subject-invariant feature space can be deployed for decoding a new subject, even with smaller training data, and improving the performance by adding other subjects' data, \ie, transfer learning.
We evaluated our proposed method using two large motor imagery EEG datasets. In addition, we analyzed our results to provide neurophysiological explanation and explicate the results from the viewpoint of transfer learning. Further, we expect that our proposed method \edit{could} be applied to other types of EEG signals.

% \section*{Acknowledgment}
% % We are very grateful to Wonsik Jung for his help on figures. 

% Can use something like this to put references on a page
% by themselves when using endfloat and the captionsoff option.
\ifCLASSOPTIONcaptionsoff
  \newpage
\fi

\bibliographystyle{IEEEtran}
\bibliography{arxiv.bib}

\end{document}